\documentclass[11pt]{article}
\usepackage[margin=1in]{geometry}
\usepackage{graphicx}
\usepackage{booktabs}
\usepackage{amsmath,amssymb}
\usepackage[colorlinks=true,linkcolor=blue,citecolor=blue,urlcolor=blue]{hyperref}
\usepackage{caption}
\title{\textbf{Depth and Scale in the Sub-150M Regime:\\ JugnuLM-53M vs JugnuLM-110M}}
\author{
  Dushyant Rajput \and Nirdesh Chauhan \and Siddharth Kosaraju \\[2pt]
  AltSlate Labs LLP \\
  \texttt{\{dushyant, nirdesh, siddharth\}@altslate.com}
}
\date{September 2026}

\begin{document}
\maketitle

\begin{abstract}
We scale our conventional sub-150M pretraining recipe from 53.5M to 109.7M parameters,
holding the method fixed (Qwen3-style decoder~\cite{qwen3} with grouped-query attention~\cite{gqa},
RoPE~\cite{rope}, SwiGLU~\cite{swiglu}, RMSNorm~\cite{rmsnorm}, QK-Norm~\cite{qknorm}, and a
z-loss~\cite{palm}; FineWeb-Edu data~\cite{fineweb}) and changing only the geometry to a
\emph{deep-and-thin} 23-layer $\times$ 576-hidden design~\cite{mobilellm}. The larger model
improves across the board---\textbf{BLiMP 78.1 $\rightarrow$ 81.3}~\cite{blimp},
\textbf{ARC-Easy 51.4 $\rightarrow$ 52.5}~\cite{arc},
\textbf{WikiText-2 byte-perplexity 2.04 $\rightarrow$ 1.95}~\cite{wikitext}---and its 81.3\% BLiMP
essentially matches GPT-X2-125M (81.28) at ${\sim}12\%$ fewer parameters. Notably the 110M model
achieves this on \emph{fewer} training tokens ($\approx$8B vs $\approx$12B), so the gain is
attributable to capacity and depth, not more data. Both models are deliberately conventional; this
report is a clean scaling control and the baseline rung (R0) of an ablation study of what further
improves models in this regime. An ablation ladder follows: value residuals~\cite{valueresidual}
(R1) and the Muon optimizer~\cite{muon} (R2) lift ARC-Easy by a cumulative $+3.6$
(52.5 $\rightarrow$ 56.1) at a near-flat BLiMP and are kept; a diverse data blend (R3) and two
logit-distillation~\cite{kd} settings (R4a/R4b) are \emph{not} kept---honest negatives. R3 pins
ARC-Easy to FineWeb-Edu's educational filtering rather than raw diversity; distillation from a 1.7B
teacher can reach the class-leading ARC-Easy (56.99 $\approx$ GPT-X2-125M) but only at a perplexity
cost that dialing KD down then erases---so R2 remains the best kept stack.
\end{abstract}

\section{Introduction}
Our earlier report introduced JugnuLM-53M, a deliberately conventional from-scratch model
competitive on the sub-150M Tiny-ML Leaderboard. A natural first question for the regime is the
cheapest and most controlled one: \emph{what does simply scaling the same recipe buy?} Here we
double the parameter count to ${\sim}110$M, change \emph{only} the geometry (deeper and narrower,
following MobileLLM's depth-over-width finding~\cite{mobilellm}), and hold every other choice fixed.
This isolates the effect of scale + depth from any change in method or data.

\section{The two models}
Identical recipe; the differences are geometry and (incidentally) token budget.

\begin{table}[h]\centering
\begin{tabular}{lrr}
\toprule
 & \textbf{JugnuLM-53M} & \textbf{JugnuLM-110M} \\
\midrule
Parameters & 53.5M & 109.7M \\
Layers & 8 & 23 \\
Hidden size & 512 & 576 \\
Heads / KV heads & 8 / 4 & 9 / 3 \\
FFN intermediate & 1792 & 1536 \\
Geometry & wide-ish & deep-thin \\
Tokens seen & $\approx$12B & $\approx$8B \\
Attention / Norm / Pos & \multicolumn{2}{c}{GQA + QK-Norm / RMSNorm / RoPE} \\
Tokenizer & \multicolumn{2}{c}{SmolLM2~\cite{smollm2} (49{,}152)} \\
\bottomrule
\end{tabular}
\caption{The two models share the entire recipe; only geometry (and token budget) differ.}
\label{tab:cfg}
\end{table}

\section{Training}
\begin{figure}[h]\centering
\includegraphics[width=0.9\textwidth]{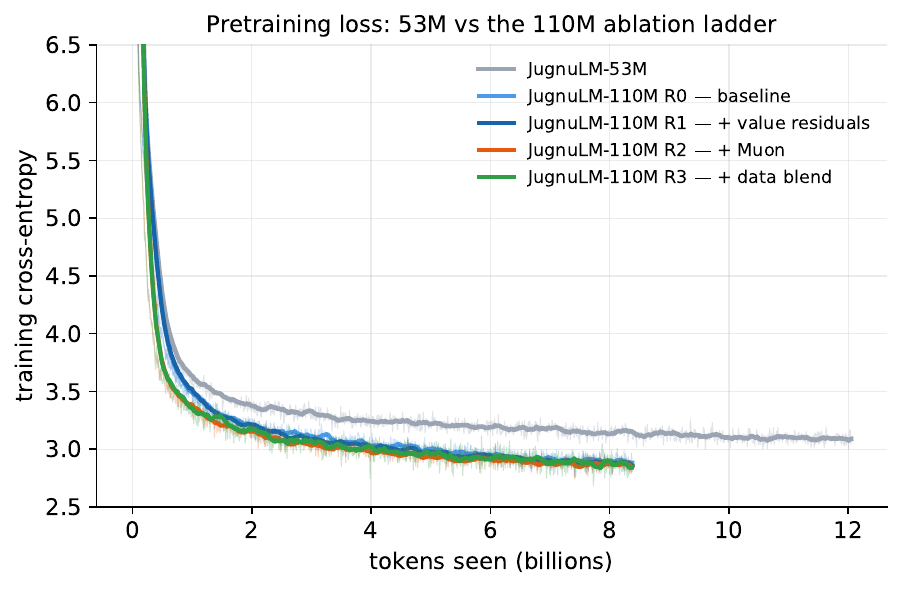}
\caption{Pretraining cross-entropy vs. tokens seen, for 53M and the three 110M rungs (R0 baseline,
R1 value residuals, R2 + Muon). All 110M variants sit below the 53M throughout and finish lower even
though they saw fewer tokens; the three 110M curves nearly coincide.}
\label{fig:loss}
\end{figure}

Both models train on FineWeb-Edu~\cite{fineweb} (sample-10BT), bf16 + DDP + \texttt{torch.compile}
on 4$\times$ RTX PRO 4500 Blackwell GPUs, with a cosine schedule (peak $1.5\times10^{-3}$ $\rightarrow$
floor $1.5\times10^{-4}$), ${\sim}0.5$M-token global batch, and a z-loss~\cite{palm} for logit
stability. The R1 and R2 rungs (below) share this training setup; R2 additionally swaps the optimizer
to Muon. Figure~\ref{fig:loss} overlays the training loss for 53M and all three 110M rungs against
tokens seen: the 110M curves track a consistently lower loss than 53M and finish lower
($\approx$2.85 vs $\approx$3.07) despite stopping at fewer tokens, while the three 110M rungs sit
close together in training loss---the levers separate them downstream, not on the raw objective
(with one exception, below).

\section{Results}
\begin{table}[h]\centering
\begin{tabular}{lrrrr}
\toprule
\textbf{Model} & \textbf{Params} & \textbf{BLiMP $\uparrow$} & \textbf{ARC-Easy $\uparrow$} & \textbf{byte-ppl $\downarrow$} \\
\midrule
Supra-50M-Instruct & 51.8M & 76.30 & 52.20 & 2.56 \\
\textbf{JugnuLM-53M} & 53.5M & 78.14 & 51.43 & 2.04 \\
\textbf{JugnuLM-110M-R0} & 109.7M & \textbf{81.25} & 52.48 & 1.95 \\
\textbf{JugnuLM-110M-R1} & 109.7M & 81.10 & 54.67 & 1.94 \\
\textbf{JugnuLM-110M-R2} & 109.7M & 80.78 & \textbf{56.10} & \textbf{1.932} \\
GPT-X2-125M & 125M & 81.28 & 57.07 & 1.86 \\
Haidass-143M & 143M & 79.05 & 60.23 & 1.89 \\
\bottomrule
\end{tabular}
\caption{All of our models against the sub-150M field. Scaling 53M $\rightarrow$ 110M (R0) lifts
every metric; BLiMP (81.25) essentially matches GPT-X2-125M (81.28) at ${\sim}12\%$ fewer parameters,
on $\approx$8B training tokens. The two ablation rungs then climb ARC-Easy to 56.10---within
${\approx}1$ point of GPT-X2-125M---while BLiMP holds. Bold marks the best of ours per column.
Comparator scores (Supra-50M-Instruct, GPT-X2-125M, Haidass-143M) are as reported on the sub-150M
Tiny-ML Leaderboard.}
\label{tab:results}
\end{table}

\begin{figure}[h]\centering
\includegraphics[width=\textwidth]{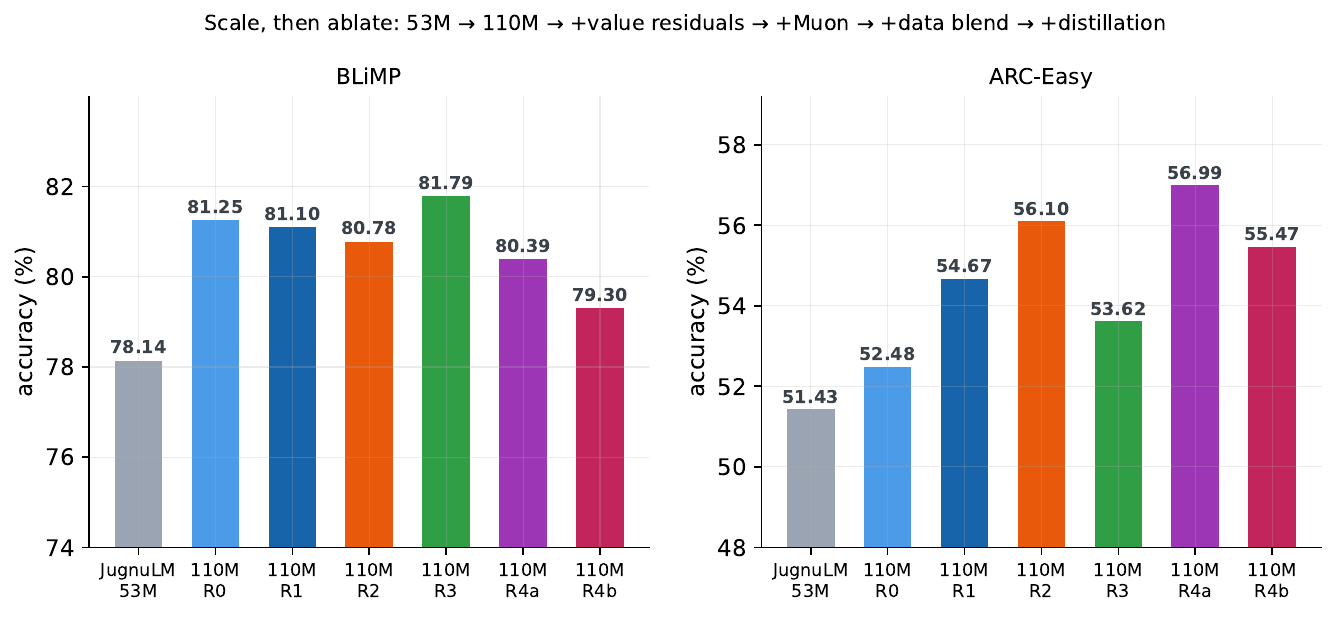}
\caption{BLiMP and ARC-Easy across all our models: 53M $\rightarrow$ 110M (R0) $\rightarrow$
+value residuals (R1) $\rightarrow$ +Muon (R2) $\rightarrow$ +data blend (R3) $\rightarrow$
+distillation (R4a heavy, R4b light). R1--R2 lift ARC-Easy and are kept. R4a posts the highest
ARC-Easy (56.99) but at a perplexity cost (see Table~\ref{tab:ablation}); R3 and R4b give ARC-Easy
back. Only R1/R2 survive the keep-if-it-beats-the-prior-rung test.}
\label{fig:bars}
\end{figure}

\begin{figure}[h]\centering
\includegraphics[width=0.97\textwidth]{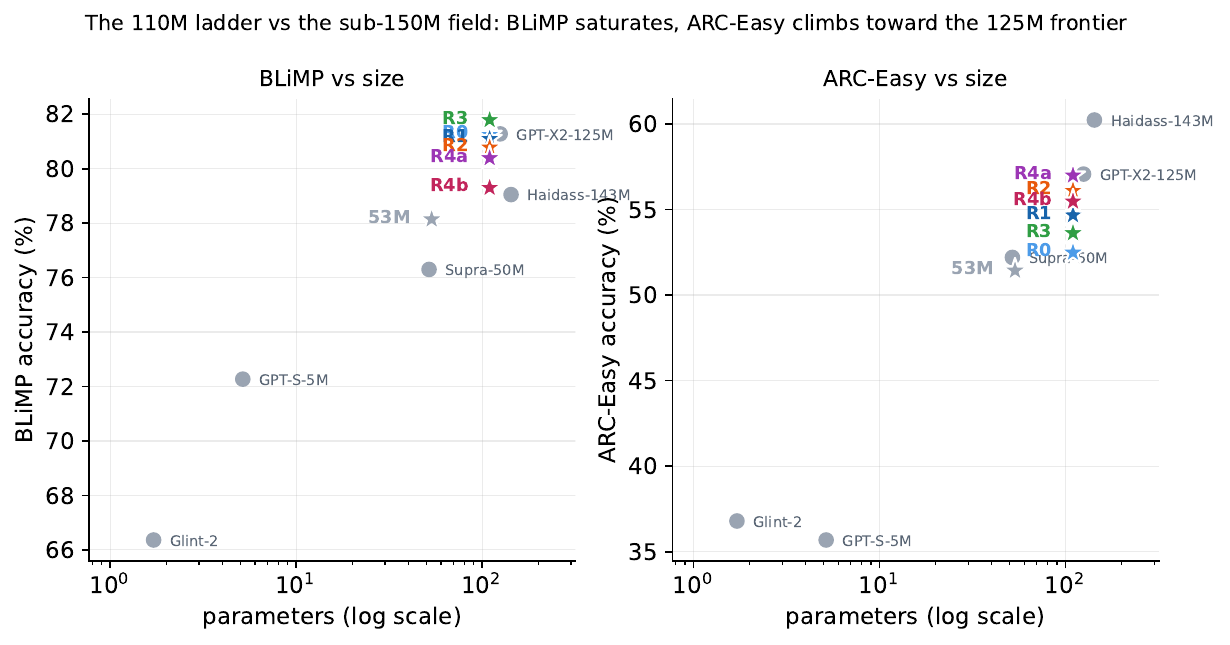}
\caption{BLiMP (left) and ARC-Easy (right) vs. parameter count (log scale), for all variants and
the sub-150M field. BLiMP saturates across the 110M rungs; ARC-Easy climbs
R0 $\rightarrow$ R1 $\rightarrow$ R2 to within ${\approx}1$ point of GPT-X2-125M, then R3 (data
blend) falls back.}
\label{fig:size}
\end{figure}

The 53M $\rightarrow$ 110M (R0) jump is $+3.11$ BLiMP, $+1.05$ ARC-Easy, and $-0.09$
byte-perplexity; the two ablation rungs then add a further $+3.6$ ARC-Easy
(Figure~\ref{fig:bars}). Figure~\ref{fig:size} places every variant against the field on both
metrics: on BLiMP-versus-size the 110M rungs cluster with the 125--143M models (BLiMP saturates),
while on ARC-Easy-versus-size the ladder climbs R0 $\rightarrow$ R1 $\rightarrow$ R2 toward the
GPT-X2-125M frontier.

\section{The ablation ladder}
With the 110M model (R0) as an honest baseline, we run an ablation ladder: add one lever at a time,
holding everything else fixed (same data, tokens, schedule, batch, seed), and keep a lever only if
it beats the previous rung. Four rungs are complete: R1 and R2 kept; R3 (data blend) and R4a/R4b
(distillation) dropped. R2 is the champion stack.

\begin{table}[h]\centering
\begin{tabular}{lrrr}
\toprule
\textbf{110M rung} & \textbf{BLiMP $\uparrow$} & \textbf{ARC-Easy $\uparrow$} & \textbf{byte-ppl $\downarrow$} \\
\midrule
R0 --- baseline & 81.25 & 52.48 & 1.95 \\
R1 --- + value residuals & 81.10 & 54.67 & 1.94 \\
R2 --- + Muon optimizer & 80.78 & \textbf{56.10} & 1.932 \\
R3 --- + data blend & \textbf{81.79} & 53.62 & \textbf{1.9092} \\
R4a --- + logit KD (heavy) & 80.39 & \textbf{56.99} & 2.178 \\
R4b --- + logit KD (light) & 79.30 & 55.47 & 1.9165 \\
\bottomrule
\end{tabular}
\caption{Each rung adds one lever; keep only what beats the prior rung on ARC-Easy (the binding
metric). R1--R2 climb ARC-Easy $+3.6$ (52.48 $\rightarrow$ 56.10) and are kept. R3 (data blend), R4a
(heavy KD) and R4b (light KD) are all \emph{not kept}---honest negatives discussed below.}
\label{tab:ablation}
\end{table}

\paragraph{R1 --- value residuals.} We add value residuals~\cite{valueresidual} (ResFormer): each
layer's value gains a learned-gated residual from the first layer's value,
$\widetilde{V}_l = V_l + \lambda_l V_1$. The learned gates converge to non-trivial values (range
$-2.5$ to $+3.2$), i.e. the residual path is genuinely used. This lifts ARC-Easy by $+2.2$ at a
BLiMP tie (SE $\approx 0.14$) and a slightly lower perplexity, so we keep it and stack the next rung
on top. (Implementation note: because value residuals are a custom attention pathway, they must be
applied at evaluation too---loading the checkpoint as a stock model silently drops the residual and
understates the result.)

\paragraph{R2 --- the Muon optimizer.} On the R1 model we swap the optimizer to Muon~\cite{muon}---
Newton-Schulz orthogonalized momentum---on the 2D hidden matrices (attention + MLP, 81.4M params),
keeping AdamW for the tied embedding/head, RMSNorm gains, and the value-residual $\lambda$ scalars
(28.3M params); both ride one shared cosine schedule (Muon peak $2\times10^{-2}$, AdamW
$1.5\times10^{-3}$). Muon lifts ARC-Easy a further $+1.43$ (54.67 $\rightarrow$ 56.10) and lowers
perplexity, at a small BLiMP dip ($-0.32$, ${\approx}2\times$ SE). Its known scaling behaviour shows
here in detail: the validation-perplexity lead over AdamW was largest early ($-26\%$ at step 1000)
and compressed to ${\approx}{-}1\%$ by the end of the fixed ${\approx}8.4$B-token budget---the
``wins the sprint, converges toward AdamW'' pattern---yet the \emph{downstream} ARC gain persisted
after the perplexity gap had closed. The optimizer helped reasoning more than final perplexity, and
we keep it.

\begin{figure}[h]\centering
\includegraphics[width=0.9\textwidth]{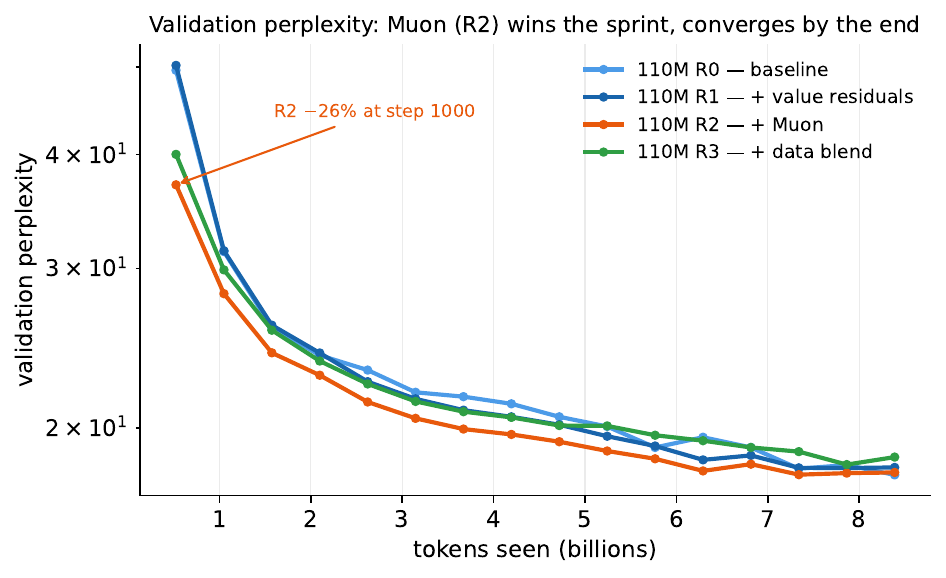}
\caption{Validation perplexity vs. tokens for the 110M rungs (fixed FineWeb-Edu probe). R0 and R1
track each other; Muon (R2) opens a large early lead ($-26\%$ at ${\approx}0.5$B tokens) that
compresses to ${\approx}{-}1\%$ by the end. R3 (blend) sits ${\approx}4$--$7\%$ higher
throughout---expected, since it trains 45\% off the pure-FineWeb-Edu probe distribution; not a
quality verdict (see R3).}
\label{fig:conv}
\end{figure}

\paragraph{R3 --- a data blend (an honest negative for ARC).} Holding architecture, optimizer,
schedule, tokens, and seed fixed at R2's, we change only the training data: instead of single-source
FineWeb-Edu, a per-sequence blend of FineWeb-Edu (55\%), DCLM-baseline (35\%), and FineMath (10\%).
The result inverts the usual ``diversity helps reasoning'' intuition. R3 gives the \emph{best} BLiMP
(81.79) and \emph{best} byte-perplexity (1.909) of any rung---broad language-modelling
improved---but ARC-Easy \emph{falls} $-2.48$ (56.10 $\rightarrow$ 53.62, ${\approx}2.5\times$ SE), a
loss on the metric that gates our rank (its efficiency score lands just below R2's). We therefore do
\textbf{not} keep R3. The likely mechanism: FineWeb-Edu's educational-quality filtering is closely
aligned with ARC-Easy's grade-school science questions, so diluting it to 55\% removes ARC-relevant
signal that the general-web (DCLM) and math (FineMath) additions do not replace. The lesson for this
regime is specific and useful: \emph{for ARC-Easy the educational data distribution matters more
than raw diversity}---the lever is more educational/science data (or distillation from a
knowledgeable teacher), not more variety.

\paragraph{R4 --- logit distillation (a two-run honest negative).} We tried offline top-16 logit
distillation~\cite{kd} from a same-tokenizer teacher (SmolLM2-1.7B, ${\approx}15\times$ the student):
the frozen teacher's top-16 next-token logits over a 6B-token region are precomputed, and the
student adds $\mathcal{L} = \alpha \cdot \text{CE} + (1-\alpha)\,\tau^2\,\text{KL}(\text{student}
\,\|\, \text{teacher})$. Two settings bracket the outcome. \textbf{R4a} (heavy KD, $\alpha{=}0.5,
\tau{=}2$, so KD ${\approx}4\times$ CE) lifts ARC-Easy to \textbf{56.99}---the best of any rung,
essentially matching GPT-X2-125M (57.07)---but the student over-imitates the teacher's softened
distribution and byte-perplexity blows up to 2.178. \textbf{R4b} (light KD, $\alpha{=}0.7,
\tau{=}1$) recovers R2-level perplexity (1.917) but \emph{loses} the ARC gain (55.47, below R2's
56.10) and BLiMP falls to 79.30. So heavy KD buys reasoning at the cost of perplexity; light KD
fixes perplexity but forfeits the reasoning---neither beats R2 on the blended efficiency score, and
\textbf{R2 remains the champion stack}. (Caveat: R4a/R4b trained on the 6B KD-logit region while R2
saw the full ${\sim}10$B pool, so the KD runs also saw less unique data; a fair retry needs
full-corpus teacher logits or a different mechanism---reverse-KL, a smaller teacher.) Offline logit
KD at this teacher--student gap and token budget is not a clean win here.

\section{Discussion}
\paragraph{The gain is from capacity + depth, not data.} The 110M model saw \emph{fewer} tokens
($\approx$8B vs $\approx$12B) yet scored higher on every metric. So this is a clean
capacity/geometry effect; a token-matched 110M would likely be stronger still. The comparison is
\emph{not} token-matched---that is a caveat, but one that only strengthens the direction.

\paragraph{Depth-thin is a sensible default here.} Following~\cite{mobilellm} we spent the extra
budget on depth (8 $\rightarrow$ 23 layers) rather than width, and the model reaches grammar
competence comparable to wider, larger, far-more-trained models.

\paragraph{It is still a conventional recipe.} Neither model uses any architectural or
methodological novelty. This report is deliberately a \emph{control}: it establishes how far plain
scaling goes, so that the ablations---value residuals~\cite{valueresidual} (R1), the Muon
optimizer~\cite{muon} (R2), a data blend (R3), and logit distillation~\cite{kd} (R4a/R4b), all
reported above---are measured against an honest, well-understood baseline (this 110M is rung R0 of
that ladder). R3 and R4 are a reminder that the ladder discipline earns its keep: a change can look
good on aggregate language modelling yet lose the metric that matters, and we drop it.

\section{Reproducibility}
Code and configs (incl. the 110M recipe, \texttt{value\_residual.py}, and \texttt{muon.py}) are
public at \url{https://github.com/AltSlate-Labs/jugnu}; weights and full model cards at
\url{https://huggingface.co/altslate} (\texttt{JugnuLM-53M}, \texttt{JugnuLM-110M} (R0), and the
ablation rungs \texttt{-R1}, \texttt{-R2}, \texttt{-R3}, \texttt{-R4a}, \texttt{-R4b}). Evaluation of
R0 uses the EleutherAI harness~\cite{lmeval}:
\begin{quote}\small\texttt{lm\_eval --model hf --model\_args pretrained=altslate/JugnuLM-110M,dtype=bfloat16 --tasks blimp,arc\_easy,wikitext --batch\_size auto}\end{quote}
The R1--R4b checkpoints add a custom value-residual attention pathway, so they load with
\texttt{trust\_remote\_code=True} and must be evaluated on the rebuilt model (loading them as a stock
Qwen3 silently drops the residual and understates the result).

\section{Conclusion}
Across a controlled study of the sub-150M regime we find that most of the gain over a 53M baseline
comes from two moves: \textbf{scaling with depth} (the deep-thin 110M matches the BLiMP of a 125M
reference at ${\sim}12\%$ fewer parameters and fewer training tokens) and \textbf{two cheap,
well-chosen levers}---value residuals (R1) and the Muon optimizer (R2)---which together lift ARC-Easy
$+3.6$ to 56.1 at a near-flat BLiMP. \textbf{JugnuLM-110M-R2 is the model we ship}: the strongest
kept stack, competitive with the top of the sub-150M field on our efficiency estimate, with ARC-Easy
the remaining gap to the leaders.

Equally informative are the rungs we \emph{dropped}. A diverse data blend (R3) improved aggregate
language modelling yet lost ARC-Easy---at this scale ARC tracks FineWeb-Edu's educational filtering,
not raw diversity. Distillation (R4a/R4b) could reach class-leading ARC-Easy (56.99 $\approx$
GPT-X2-125M) but only at a perplexity cost that a gentler setting then erased; offline logit KD from
a 1.7B teacher was not a clean win at this teacher--student gap and token budget. The ladder
discipline---keep a change only if it beats the previous rung on the metric that matters---is what
turns these into results rather than noise.

Two directions follow. First, a \textbf{final competitive run} on the winning stack at a larger token
budget ($\approx$25--40B) with a warmup-stable-decay schedule and decay-phase upweighting of
educational data---the levers most likely to close the ARC gap. Second, distillation remains the
highest-ceiling lever and deserves a better mechanism (reverse-KL, a smaller teacher to shrink the
capacity gap, or full-corpus teacher logits) before it is dismissed.

\bibliographystyle{unsrt}
\bibliography{refs}

@article{fineweb,
  title={The FineWeb Datasets: Decanting the Web for the Finest Text Data at Scale},
  author={Penedo, Guilherme and Kydl{\'\i}{\v{c}}ek, Hynek and others},
  journal={arXiv preprint arXiv:2406.17557}, year={2024}
}

@article{smollm2,
  title={SmolLM2: When Smol Goes Big -- Data-Centric Training of a Small Language Model},
  author={Allal, Loubna Ben and Lozhkov, Anton and others},
  journal={arXiv preprint arXiv:2502.02737}, year={2025}
}

@article{qwen3,
  title={Qwen3 Technical Report},
  author={Yang, An and others},
  journal={arXiv preprint arXiv:2505.09388}, year={2025}
}

@article{gqa,
  title={GQA: Training Generalized Multi-Query Transformer Models from Multi-Head Checkpoints},
  author={Ainslie, Joshua and Lee-Thorp, James and others},
  journal={arXiv preprint arXiv:2305.13245}, year={2023}
}

@article{rope,
  title={RoFormer: Enhanced Transformer with Rotary Position Embedding},
  author={Su, Jianlin and Lu, Yu and others},
  journal={arXiv preprint arXiv:2104.09864}, year={2021}
}

@article{swiglu,
  title={GLU Variants Improve Transformer},
  author={Shazeer, Noam},
  journal={arXiv preprint arXiv:2002.05202}, year={2020}
}

@article{rmsnorm,
  title={Root Mean Square Layer Normalization},
  author={Zhang, Biao and Sennrich, Rico},
  journal={arXiv preprint arXiv:1910.07467}, year={2019}
}

@article{qknorm,
  title={Query-Key Normalization for Transformers},
  author={Henry, Alex and Dachapally, Prudhvi Raj and others},
  journal={arXiv preprint arXiv:2010.04245}, year={2020}
}

@article{palm,
  title={PaLM: Scaling Language Modeling with Pathways},
  author={Chowdhery, Aakanksha and others},
  journal={arXiv preprint arXiv:2204.02311}, year={2022}
}

@article{mobilellm,
  title={MobileLLM: Optimizing Sub-billion Parameter Language Models for On-Device Use Cases},
  author={Liu, Zechun and Zhao, Changsheng and others},
  journal={arXiv preprint arXiv:2402.14905}, year={2024}
}

@article{blimp,
  title={BLiMP: The Benchmark of Linguistic Minimal Pairs for English},
  author={Warstadt, Alex and Parrish, Alicia and others},
  journal={Transactions of the Association for Computational Linguistics}, year={2020}
}

@article{arc,
  title={Think You Have Solved Question Answering? Try ARC, the AI2 Reasoning Challenge},
  author={Clark, Peter and Cowhey, Isaac and others},
  journal={arXiv preprint arXiv:1803.05457}, year={2018}
}

@article{wikitext,
  title={Pointer Sentinel Mixture Models},
  author={Merity, Stephen and Xiong, Caiming and others},
  journal={arXiv preprint arXiv:1609.07843}, year={2016}
}

@misc{lmeval,
  title={The Language Model Evaluation Harness},
  author={Gao, Leo and Tow, Jonathan and Abbasi, Baber and Biderman, Stella and others},
  year={2024}, month={7}, publisher={Zenodo}, version={v0.4.3},
  doi={10.5281/zenodo.12608602}, url={https://doi.org/10.5281/zenodo.12608602}
}

@article{valueresidual,
  title={Value Residual Learning},
  author={Zhou, Zhanchao and Wu, Tianyi and Jiang, Zhiyun and Obeid, Fares and Lan, Zhenzhong},
  journal={arXiv preprint arXiv:2410.17897}, year={2024}
}

@misc{muon,
  title={Muon: An Optimizer for the Hidden Layers of Neural Networks},
  author={Jordan, Keller and others}, year={2024},
  howpublished={\url{https://github.com/KellerJordan/modded-nanogpt}}
}

@article{kd,
  title={Distilling the Knowledge in a Neural Network},
  author={Hinton, Geoffrey and Vinyals, Oriol and Dean, Jeff},
  journal={arXiv preprint arXiv:1503.02531}, year={2015}
}
\end{document}